\documentclass[14pt]{report}

\usepackage{amsmath}
\usepackage{amsfonts}
\usepackage{amssymb}
\usepackage{graphicx}
\usepackage{hyperref}
\usepackage{abstract}
\usepackage{caption}
\usepackage{mathptmx}
\usepackage{acronym}
\usepackage{todonotes}
\usepackage{ragged2e}
\usepackage{listings}
\usepackage{xkeyval}
\usepackage[numbers, square]{natbib}
\usepackage{array}
\usepackage{etoolbox} 
\newcolumntype{P}[1]{>{\centering\arraybackslash}p{#1}}
\usepackage{rotating} 
\usepackage{graphicx} 

\makeatletter
\patchcmd{\lst@MakeCaption}{\normalfont}{\footnotesize}{}{}
\makeatother

\lstdefinestyle{mystyle}{
    backgroundcolor=\color{lightgray},
    basicstyle=\ttfamily\small, 
    keywordstyle=\bfseries\color{blue},
    commentstyle=\itshape\color{gray},
    stringstyle=\color{red},
    numbers=left, 
    numberstyle=\tiny\color{gray},
    breaklines=true, 
    breakatwhitespace=true,
    frame=single,
}

\lstnewenvironment{mycode}[1][]%
  {\lstset{language=Python,basicstyle=\small\ttfamily,#1}}%
  {}

\begin{document}

\title{Prompt-Efficient Fine-Tuning for GPT-like Deep Models to Reduce Hallucination and to Improve Reproducibility in Scientific Text Generation Using Stochastic Optimisation Techniques}
\author{Authors: Daniil Sulimov}
\date{}

\maketitle
\thispagestyle{empty}

\setcounter{page}{2}
\tableofcontents

\chapter*{Annotation}
Large Language Models (LLMs) have demonstrated impressive performance in a variety of language-related tasks, including text generation, machine translation, text summarising. Sometimes the result produced by a LLM turns out to be inaccurate. This thesis aims to fine-tune the existing LLM, GPT-2 by OpenAI, to reduce model's hallucinations and increase the answers' reproducibility in mass spectrometry. 

The research involved the application of the following scope of skills: data engineering, stochastic modelling, data science and statistics. I used two servers for all experiments: cHARISMa Higher School of Economics (HSE) server for fine-tuning and AI for Computational biology (AIC) server, where I run Docker images, necessary for the data preprocessing.

Our fine-tuned model was named MassSpecGPT (MS-GPT). The thesis includes the novel approach of reproducibility score computations and calculation of Wilcoxon rank sum statistical test to compare the fine-tuned model MS-GPT against the base GPT-2 by OpenAI in reproducibility domain. The selection of optimal parameters (optimizer, learning rate) was based on several factors: validation error, run time, random-access memory (RAM) usage and Electricity usage. The fine-tuning of the model involved Low-Rank Adaptation of Large Language Models (LoRA) adapters, the state-of-the art (SOTA) method by now.
I used common Natural Language Generation (NLG) evaluation metrics to compare the models' accuracies: Bilingual Evaluation Understudy (BLEU), Recall-Oriented Understudy for Gisting Evaluation (ROUGE) and Perplexity. 

As the result of the research, the BLEU score increased from 0.33 to 0.34, ROUGE-1 - from 0.42 to 0.44, ROUGE-L - from 0.57 to 0.62, Perplexity reduced from 13586.37 to 10092.12 and reproducibility score went from 0.83 to 0.84. Statistically significant under 5\% significance level turned out to be Perplexity score and reproducubility.  

\chapter*{Acronyms and Abbreviations}
\addcontentsline{toc}{chapter}{Acronyms and Abbreviations}

\begin{acronym}
  \item[LLM] Large Language Model
  \item[LM] Language Model
  \item[LoRA] Low-Rank Adaptation
  \item[GPT] Generative Pre-trained Transformer
  \item[NLP] Natural Language Processing
  \item[MS] Mass Spectrometry
  \item[RNN] Recurrent Neural Network
  \item[LSTM] Long-Short Term Memory
\end{acronym}

\chapter{Introduction}

\section{Background}
A language model (LM) serves as a probabilistic framework for understanding natural language. It assesses the likelihood of various linguistic elements, including symbols, tokens, and sequences of tokens, within a given context \citep{voita2020nlpCourse}. Specifically, a large language model (LLM) like the one described here is capable of generating coherent sequences of words, forming sentences or paragraphs. Over time, the field has evolved through various approaches, among which are Recurrent Neural Networks (RNNs) \citep{Rumelhart1986LearningIR, schmidt2019recurrent}, Long-Short Term Memory Networks (LSTMs) \citep{vennerød2021long, 10.1162/neco.1997.9.8.1735}, and Generative Pre-Trained Transformers (GPT) \citep{yenduri2023generative}. These advancements have significantly enhanced the model's ability to comprehend and generate human-like text.

Considerable progress has been made in the improvement of text generating LLMs over past years, which reflects in reaching significant results on such benchmarks datasets for Language Modelling as WikiText-103\citep{grave2016improving}, WikiText2, Text8, etc. Constant improvements of the text generated quality are presented almost monthly, showing how significant the growth of the sphere has become. These enhancements have been made by developing the LLM itself and then fine-tuning\citep{tian2023finetuning} it on the specific tasks. Apart from the achieved results, over the years, we observe how LLMs grow in terms of size of parameters. This can be seen in the Figure\ref{fig:llm_sizes}. 

\begin{figure}[h]\label{fig:llm_sizes}
    \centering
    \includegraphics[width=1\textwidth]{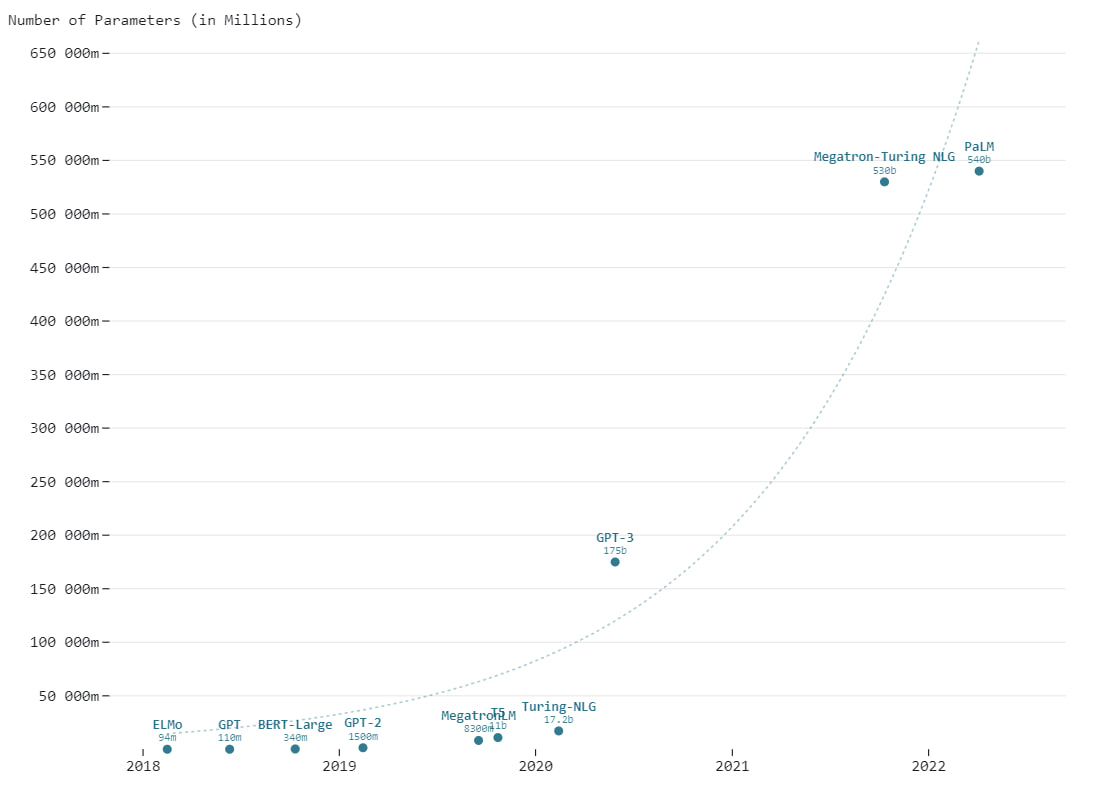}
    \caption{LLMs' sizes evolution.}
    \label{fig:llm_sizes}
    The picture was taken from \href{https://observablehq.com/@sorami/sizes-of-large-language-models}{observablehq.com}.
\end{figure}

Within 4 years from 2018, the size of models has increased for near 6000 times, and consequently, the amount of data needed either for training or fine-tuning LLMs has risen drastically. Basically, it reflects several points: the interest towards such models has grown exponentially, the usage of LLMs has spread all around, encompassing the majority of life spheres. Indeed, nowadays people rely increasingly on LLMs, especially after the release of Chat-GPT\footnote{https://openai.com/blog/chatgpt/} by OpenAI.

However, LLMs are facing several challenges. First of all, while training, each LLM is being given huge amount of texts from the various Internet web pages and quite often these materials are biased in one or another manner. It means, that when a user wants, for example, generate some essay, he receives the text that is likely to reflect the point of view of some specific society group.

Another risk in LLMs is hallucinations\citep{xiao2021hallucination}. Hallucinations are the piece of information, incorrectly generated by LLM but pretending to be true one. They have become one of the stand-out problems in LLM development\citep{10.1145/3571730} for a while. For example, the risk of hallucinations is crucial when a user tries to get some piece of knowledge in the medical field. 

Our approach aims to tackle these issues and change the behaviour of LLMs to more reproducible and accurate in computational mass spectrometry field.

\section{Research Objectives}
The training of LLMs with vast amounts of data from the Internet, e.g. the total amount of data used for GPT-3 models training is 300 billion words\citep{brown2020language}, inevitably leads to worse reproducibility and accuracy of the answers. This problem results in several issues:
\begin{itemize}
    \item Misinformation and disinformation that a user gets through relying on the LLMs. 
    \item Also, in such cases some bias can occur, where the model outputs the information that can be potentially discriminatory.
\end{itemize}
Especially, these problems turn out to be crucial when someone uses LLM in the context of educational source. Hence, getting a incorrect piece of information that, may outcome in severe consequences, such as utilising it in the further usage. Moreover, these wrong data can go public by spreading.

In order to prevent such situations and help researchers in the field of computational mass spectrometry to get the checked and verified data from the scientifically proven prospective, we set our objective to improve existing LLMs for more accurate and reproducible answers.

\section{Improving the Accuracy of LLMs}
\label{review}
In the up-to-date literature, we can encounter the studying of the LLM's tuning techniques from several points of view. First, in terms of LLM development one of the core places takes the optimisation technique, used during training process. Among the set of possible function to pick some of them are in the spotlight of almost every research, e.g. SGD (Stochastic Gradient Descent)\citep{schoenauersebag2017stochastic}, Adagrad (Adaptive Gradient)\citep{JMLR:v12:duchi11a}, RMSProp, Adam (Adagrad+RMSProp)\citep{kingma2017adam}.

Second, prompt engineering process itself encompasses several methods that could be split into two groups:
\begin{itemize}
    \item With modifying the parameters of initial model
    \item Without modifying the initial weights
\end{itemize}
The latter includes such approaches like P-Tuning, Prefix-Tuning and the former includes Fine-Tuning, encompassing Parameter-Efficient Fine-Tuning (PEFT).

P-Tuning is a method to tune the initial model with trainable embeddings represented as ${(x_{i}, y_{i})}_{i}$ - labelled dataset and conducting further back-propagation to minimise the loss function\citep{liu2023gpt}. Prefix-Tuning prepends a sequence of continuous task-specific vectors to the input, which is called a prefix and only these vectors are optimised by the model while training\citep{li2021prefixtuning}.

\begin{figure}[h]
    \centering
    \includegraphics[width=1.0\textwidth]{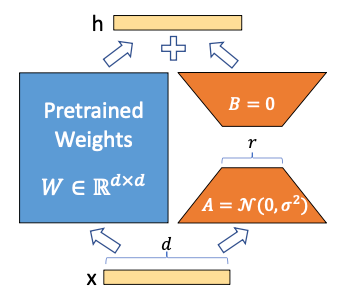}
    \caption{Illustration of LoRA adapters mechanism.}
    \label{fig:lora-mechanism}
    The picture was taken from \href{https://medium.com/@yujingda01/lora-from-a-gradient-perspective-an-overview-analysis-speculation-and-extension-a8c3249af8d8}{\url{medium.com}}.
\end{figure}

Fine-tuning is a technique that is employed to either update the initial weights of the model with the new data or train some additional layers. This method requires updating and storing all the parameters of the LLM\citep{li2021prefixtuning}. Laying on the up-to-date models' sizes, this approach could be computational inefficient. To mitigate this issue, LoRA (Low-Rank Adaptation) adapters have been proposed. The main idea is presented in the Figure \ref{fig:lora-mechanism}.

LoRA adapters represent two matrices $A$ and $B$, which consist of trainable weights, whereas model's initial pretrained weights stay immutable during training. Once the weights in $A$ and $B$ are optimised, matrices are multiplied by each other to fit the matrix dimension of initial weights. Therefore, by using LoRA we only need
to store and load a small number of task-specific parameters in addition to the pre-trained model for each task, greatly boosting the
operational efficiency when deployed\citep{hu2021lora}.

Since the development of the first LLMs, they have diffused in various spheres, including computational biology as well. In particular, one of their applications could be decoding the language embedded in DNA, e.g. universal genetic code, elucidating the translation of
DNA into proteins, has recently garnered attention for deciphering biological functions through
models based on BERT or GPT architectures, producing such models as DNABERT\citep{10.1093/bioinformatics/btab083}. Another model that has been developed recently is DNAGPT, a generalised pre-trained model for DNA sequences\citep{zhang2023dnagpt}.

Therefore, there are several ways how to adapt existing LLM to particular needs in different domains, including computational biology.

\chapter{Methodology}

\section{Data Collection and Preprocessing}
In total we managed to download several thousands of open-access research articles related to computational mass spectrometry. This was facilitated by securing an API key from the Semantic Scholar research database\footnote{https://www.semanticscholar.org/product/api}, which enabled us to comprehensively access and retrieve relevant papers. Subsequent to acquisition, we have developed an algorithm tailored to automate the downloading procedure. This script was executed on the AIC-lab (Laboratory on AI for Computational Biology) server, operating in the server’s background. The code which downloaded the articles is available in the Appendix 4.1. 

The next step was to convert the articles from PDF format to plain text format. We tested the following converting approaches: in-built Python libraries for preprocessing PDFs (PyPDF4\footnote{https://pypi.org/project/PyPDF4/}, PyMuPDF\footnote{https://pypi.org/project/PyMuPDF/}, PDFMiner\footnote{https://pypi.org/project/pdfminer/}), GROBID\citep{grobid}. The first three programs have shown approximately the same accuracy, having the preprocessing of two-columned documents as the main drawback. The key disadvantage was that in-built libraries consider lines in two-columns as one and concatenate them. Hence, among the aforementioned methods the GROBID stood out to be the best open-source working software program. 
	GROBID converts PDF documents into XML/TEI encoded documents that are structured and tailored for technical and scientific texts. GROBID is based on a deep learning model implemented through the DeLFT library\citep{DeLFT} for processing text for tasks like sequence labeling and classification, implemented in Python environment through the Java Embedded Python (JEP)\footnote{https://github.com/ninia/jep} interface and run in a  Docker environment on AIC server.
	
    I completed Python scripts to extract the scientific text from the XML/IEI files. The extraneous elements such as headers, footers, and references were removed. The code responsible for the mentioned is available in the Appendix 4.2.  The resulting corpus of data was then critically reviewed to ensure its suitability for the subsequent stages of our computational analysis. The final total amount of the ready-to-train data exceeds 2.5 Giga Bytes and was uploaded to the HSE cHARISMa cluster\citep{charisma}.

In summary, the chosen methodology for PDF preprocessing was a careful blend of existing technology and modern deep learning innovations, crafted to meet the demands of our research goals efficiently.

\section{Fine-tuning the LLM} \label{fine-tuning}
After paper downloading and text extraction, the fine-tuning approach was designed in the following way:
\begin{itemize}
    \item Choosing the LLM to fine-tune;
    \item Selecting the range of optimizers and learning rates - parameters that are responsible for training speed, for research;
    \item Sampling the text corpus for the grid search needs;
    \item Equip the initial model with LoRA adapters;
    \item Conduct grid-search on the sampled data to pick the best performing optimiser and learning rate;
    \item Launch the training process on the whole text corpus with the best model parameters
\end{itemize}
As the choice of the base model to fine-tune, I selected the most downloaded one (21 mln. of downloads) from the Open Hugging Face AI Community\citep{wolf2020huggingfaces}, i.e. GPT-2, developed by the OpenAI\citep{radford2019language}. There are three open-source versions of GPT-2 model: small, medium and large. They differ from each other in their sizes, i.e. GPT-2-small contains 124 million of parameters, medium - 355 million, large - 774 million. For this project we selected small version of GPT-2 due to limitations in computational power. Afterwards, the initial model was equipped with LoRA adapters, using the PEFT library\citep{peft}, it allowed to extend the GPT-2 model with more trainable weights for better performance.
In terms of optimizers, the ones that were named in Section \ref{review}, were selected, i.e. SGD, AdaGrad, RMSProp, Adam with the following learning rates: $10^{-2}$, $5*10^{-3}$, $5*10^{-4}$. From the full text corpus, obtained after the preprocessing of PDF files, we picked $10\%$ of data, that was further used in the finding the most advantageous optimization parameters. The code of the grid-search procedure is available in the Appendix 4.3.

Therefore, the model training of the full text corpus was conducted after obtaining the optimal parameters, described above.

Also, let us state some points essential for the fine-tuning process:
\begin{itemize}
    \item Tokenization - a process of breaking down the text into smaller manageable pieces, i.e. words. Also a tokenizer is responsible for converting text data into numerical representation.
    \item TextDataset is used for loading text, its tokenization and breaking into parts with the parameter block\_size, specifying the maximum length of the tokenized sequences.
    \item Data Collator is responsible for preprocessing data in batches and padding the sequences. Padding is a technique used in natural language processing (NLP) and deep learning to ensure that input sequences are of uniform length. The essence of padding technique is in adding special characters to the shorter sequence.
    \item Gradient accumulation steps - the number of steps being accumulated before the parameters update.
    \item Weight decay - a regularisation technique used to penalise the large weights to prevent overfitting.
    \item Batch size - the number of training examples utilised in one iteration of the training process.
\end{itemize}
The aforementioned terms define the training process.

\subsection{Mathematical properties of the model training}
Let us explain some mathematical properties regarding the optimizers we used during the grid-search.

SGD:
\begin{equation}
    \theta^{(t+1)} = \theta^{(t)} - \eta \cdot \text{SGD}(\nabla_{\theta} \mathcal{L}(\theta^{(t)}))
\end{equation}, where:
\begin{itemize}
    \item \(\theta^{(t)}\) \text{: Model parameters at iteration t}
    \item \(\eta\) \text{: Learning rate, controlling the step size in the parameter space.}
    \item \(\nabla_{\theta}\) \(\mathcal{L}(\theta^{(t)}\))\text{: Gradient of the loss function with respect to the model parameters at iteration t.}
    \item \text{SGD}\(\nabla_{\theta} \mathcal{L}(\theta^{(t)})\) \text{: Stochastic Gradient Descent, using stochastic estimates of the gradient for updating } \(\theta\).
\end{itemize}

Adagrad:
\begin{equation}
G^{(t+1)} = G^{(t)} + (\nabla_{\theta} \mathcal{L}(\theta^{(t)}))^2
\end{equation}

\begin{equation}
\theta^{(t+1)} = \theta^{(t)} - \frac{\eta}{\sqrt{G^{(t+1)}} + \epsilon} \cdot \nabla_{\theta} \mathcal{L}(\theta^{(t)})
\end{equation}, where:
\begin{itemize}
    \item \(G^{(t)}\) \text{: The sum of squared gradients up to iteration t.}
    \item \(\epsilon\) \text{: A small constant to prevent division by zero.}
\end{itemize}

RMSProp:
\begin{equation}
G^{(t+1)} = \beta \cdot G^{(t)} + (1 - \beta) \cdot (\nabla_{\theta} \mathcal{L}(\theta^{(t)}))^2
\end{equation}

\begin{equation}
\theta^{(t+1)} = \theta^{(t)} - \frac{\eta}{\sqrt{G^{(t+1)}} + \epsilon} \cdot \nabla_{\theta} \mathcal{L}(\theta^{(t)})
\end{equation}, where:
\begin{itemize}
    \item \(G^{(t)}\) \text{: The moving average of squared gradients.}
    \item \(\beta\) \text{: A decay factor for the moving average.}
    \item \(\epsilon\) \text{: A small constant to prevent division by zero.}
\end{itemize}

Adam:
\begin{equation}
    m^{(t+1)} = \beta_1 \cdot m^{(t)} + (1 - \beta_1) \cdot \nabla_{\theta} \mathcal{L}(\theta^{(t)})
\end{equation}
\begin{equation}
    v^{(t+1)} = \beta_2 \cdot v^{(t)} + (1 - \beta_2) \cdot (\nabla_{\theta} \mathcal{L}(\theta^{(t)}))^2
\end{equation}
\begin{equation}
    \theta^{(t+1)} = \theta^{(t)} - \eta \cdot \frac{m^{(t+1)}}{\sqrt{v^{(t+1)}} + \epsilon}
\end{equation}, where:
\begin{itemize}
    \item \(m^{(t)}\) \text{: First moment estimate of the gradient.}
    \item \(v^{(t)}\) \text{: Second moment estimate of the gradient.}
    \item \(\beta_1\) \text{: Exponential decay rate for the first moment estimate.}
    \item \(\beta_2\) \text{: Exponential decay rate for the second moment estimate.}
    \item \(\epsilon\) \text{: Small constant to prevent division by zero.}
\end{itemize}
The GPT-2 model works on conditional generation. Let \( G \) represent the model. Given an input string \( I \), the model generates an output \(  O\):

\begin{equation}
     O = G(I)
\end{equation}

The generation process seeks to optimize the model parameters \( \theta_G \) to maximize the probability of generating the correct query. This is expressed as follows:

\begin{equation}
    \theta_G = \arg \max_{\theta_G} \log P(Q |\theta_G) .
\end{equation}

\begin{itemize}
    \item \( \theta_G \): Model parameters of the GPT-2 model.
    \item \( \log P(Q | I; \theta_G) \): The logarithm of the conditional probability of a correct generated text \( Q \) given input \( I \) with model parameters \( \theta_G\) .
\end{itemize}

Optimisers are responsible for iterative parameter update of the model to achieve the minimisation of the negative log likelihood loss function, which is set by default in GPT-2:
\begin{equation}
    NLL(y) = -{\log(p(y))}
\end{equation}
\begin{equation}
    \min_{\theta} \sum_y {-\log(p(y;\theta))}
\end{equation}
\begin{equation}
    \max_{\theta} \prod_y p(y;\theta)
\end{equation}

Mathematically, the aim of applied LoRA adapters is described the following way:
\begin{equation}
     O = G(I)
\end{equation}

The generation process seeks to optimise the adapter parameters \( \theta_A \) to maximise the probability of generating the correct text. This is expressed as follows:

\begin{equation}
    \theta_A = \arg \max_{(\theta_A)} \log P(O | \theta_G, \theta_A) .
\end{equation}

\begin{itemize}
    \item \( \theta_G \): Model parameters of the initial model.
    \item \( \theta_A \): Adapter parameters of the LoRA adapter.
    \item \( \log P(O | I; \theta_G, \theta_A) \): The logarithm of the conditional probability of a correctly generated text \( O \) given input \( I \) with \( \theta_G \) and \( \theta_A \) .
\end{itemize}

\section{Model architecture}
The architecture of the GPT-2 model is presented in the Figure 2.1.
\begin{figure}[h]
    \centering
    \includegraphics[width=1.0\textwidth]{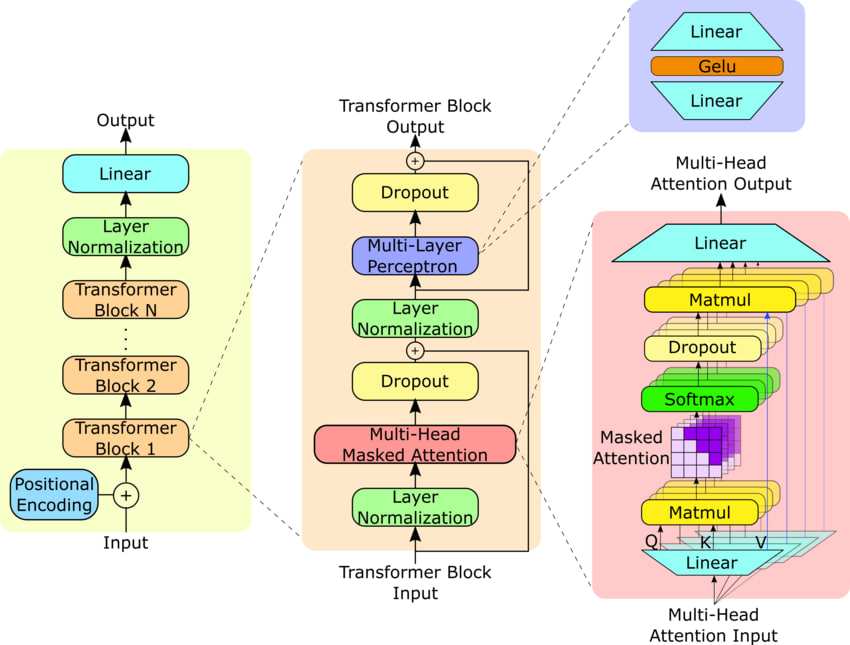}
    \label{fig:gpt2}
    \caption{GPT-2 architecture}
    The picture was taken from \citep{gpt2archit}.
\end{figure}

GPT-2 is a decoder-only model, developed for auto regressive tasks, e.g. text generation. Each decoder is presented as a transformer block. Since we use GPT-2-small version, in our case N = 12. The principle of a decoder block logic is based on masked self-attention. Attention technique itself is illustrated in the Figure 2.2 and the computations made during self-attention, presented in the Figure 2.3 are practically the same:
\begin{itemize}
    \item Each word $x$ has its vector called "Query" - $Q_{x}$;
    \item It is multiplied by the "Key" ($K_{i}$) vector of every word in the sequence except $x$;
    \item This score is used in softmax function: $\sigma(z_i) = \frac{e^{z_{i}}}{\sum_{j=1}^K e^{z_{j}}} \ \ \ for\ i=1,2,\dots,K$;
    \item The probability, obtained after the softmax application, is multiplied by the corresponding "Value" ($V_{i}$) vector;
    \item $Attention(Q_{x}, K_{i}, V_{i})=softmax(\frac{Q_{x}*K_{i}^T}{\sqrt{d_{k}}})*V_{i}$, where $d_{k}$ is a vector of the dimensionality $K_{i}$, $V_{i}$, used as a normalisation.  
\end{itemize}
\begin{figure}[h]
    \centering
    \includegraphics[width=1.0\textwidth]{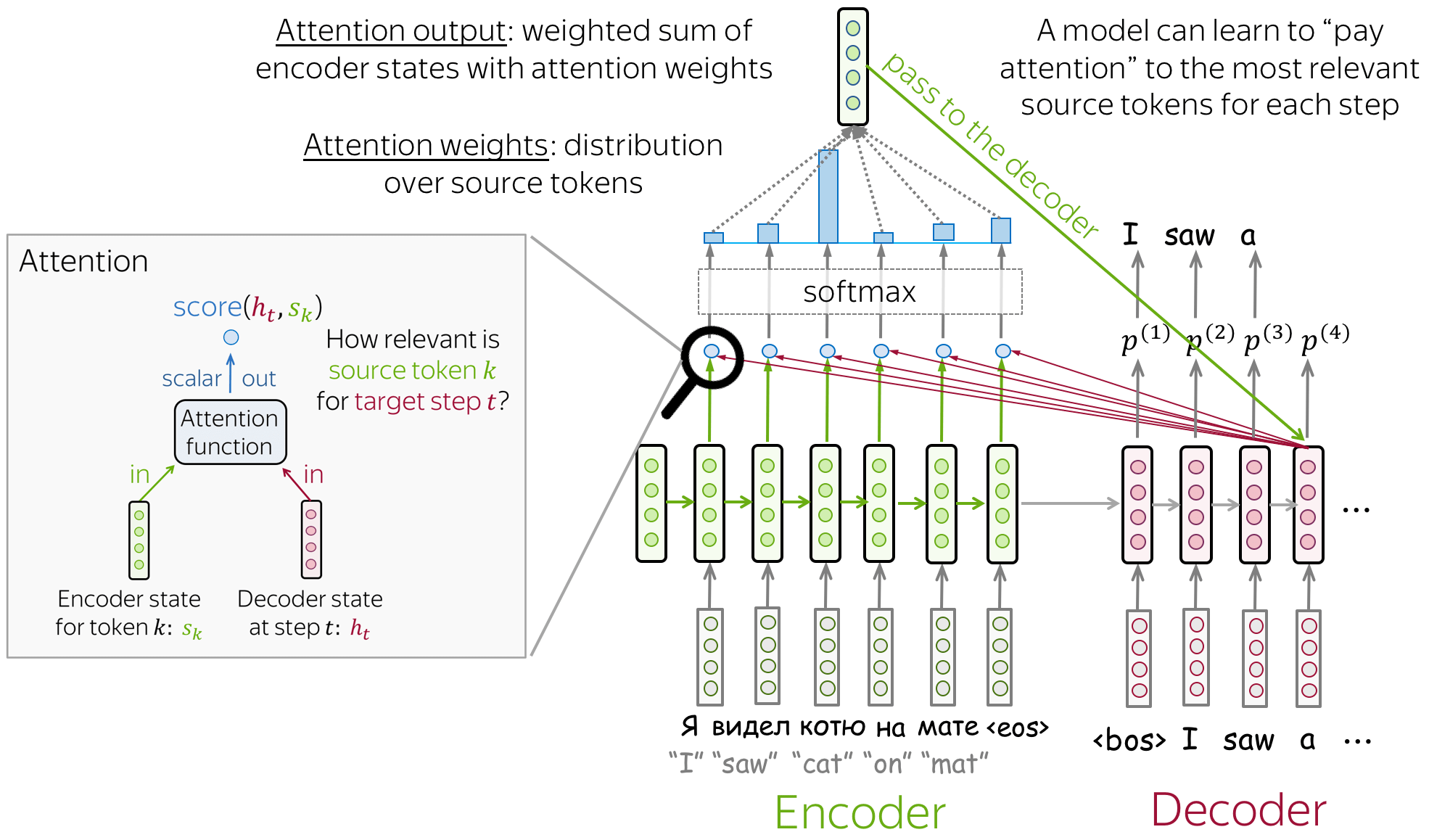}
    \label{fig:attention}
    \caption{Attention mechanism.}
    The picture was taken from \href{https://lena-voita.github.io/nlp_course/seq2seq_and_attention.html#attention_intro}{\url{lena-voita.github.io}}.
\end{figure}

\begin{figure}[h]
    \centering
    \includegraphics[width=1\textwidth]{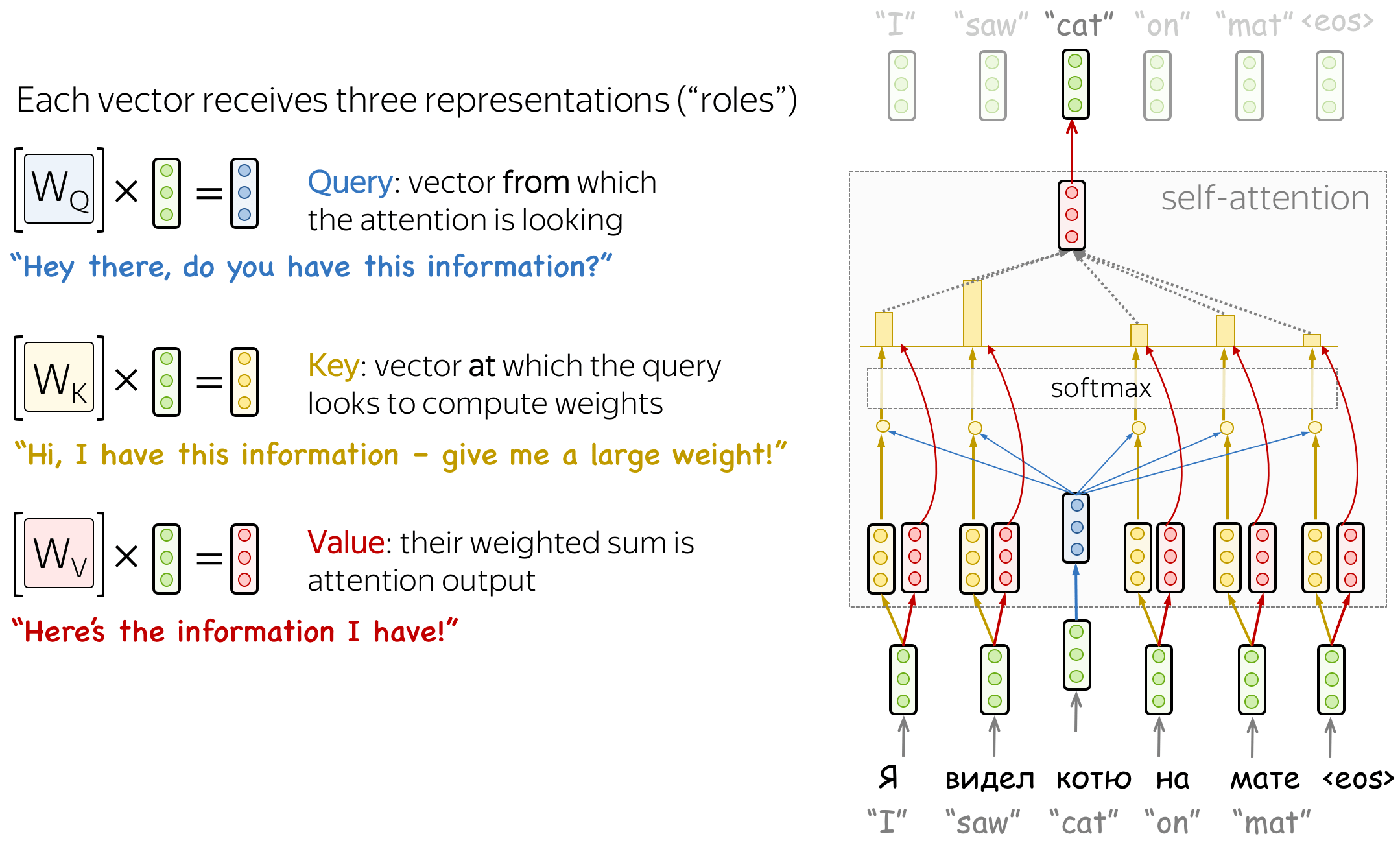}
    \label{fig:qkv}
    \caption{Self-Attention computations.}
    The picture was taken from \href{https://lena-voita.github.io/nlp_course/seq2seq_and_attention.html#attention_intro:~:text=weights%20to%20this-,token).,-The%20formula%20for}{\url{lena-voita.github.io}}.
\end{figure}
The difference between masked and unmasked self-attention is that in the masked a word can not get values and keys from the following words, only preceding\citep{vaswani2023attention}. Attention mechanism can be interpreted as the "communication between the words" to find out the relation between them and updating their embeddings, i.e. vector representations. 

Each attention head, i.e. decoder block has its own query, key and value matrices(c\_attn). Hence, after the last decoder layer, for each word in the sequence, its resulting vector is the concatenation of all (12 in our case) obtained vectors. The next step is the multiplication of the concatenated vector by the weight matrix (c\_proj) that projects the results of the attention heads into the output vector of the self-attention sublayer. Having completed this step, the vector is sent to the multi-layer perceptron (mlp) and after proceeding throughout all 12 transformer blocks we get the output. Finally, this output is passed to lm\_head after normalisation in order to get the next predicted word.

LoRA adapters, described above with r = 4, are applied to the following layers: c\_attn, c\_proj and lm\_head. The overall number of parameters is 124.410.624, among them there are 4.876.832 trainable. i.e. 3.92\%. Detailed presentation of the initial weights is presented in the Figure 2.4.
\begin{figure}[h]
    \centering
    \includegraphics[width=1\textwidth]{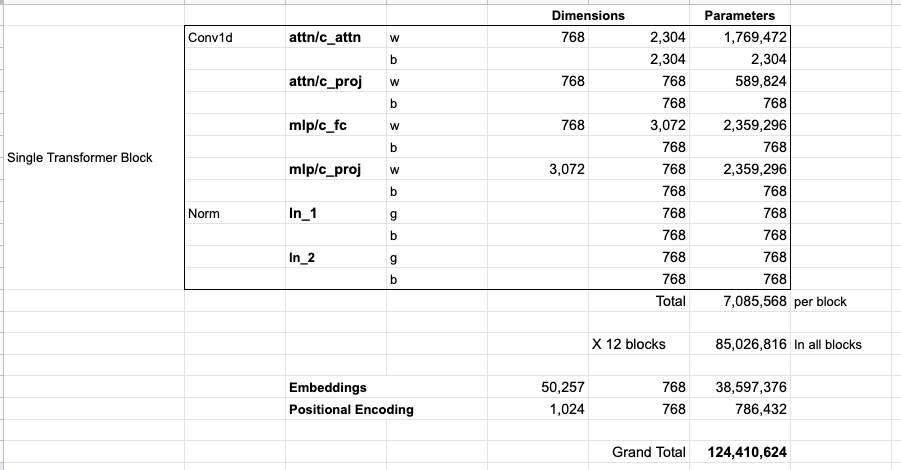}
    \label{fig:gpt2_weights}
    \caption{GPT-2 weights.}
    The picture was taken from \href{https://jalammar.github.io/illustrated-gpt2/#part-2-illustrated-self-attention:~:text=model%2C%20then%20I-,have,-tallied%20them%20here}{\url{jalammar.github.io}}.
\end{figure}

\section{Results} \label{evaluation}
\subsection{Parameter Selection}
As it was stated in the Section \ref{fine-tuning}, we trained the model on the sample from the full text corpus to find out the best learning rate and the optimiser among selected. We employed Google Colab\footnote{https://colab.research.google.com/} server with NVIDIA GPU (Graphics processing unit), that accelerates neural network training, provided by free with the limits. We chose several features as the criteria for all combinations learning rate - optimiser: 
\begin{itemize}
    \item Validation error
    \item Running time, sec
    \item Memory usage, MB
    \item Electricity usage, kWh (which can be interpreted as carbon emission and reflects environmental impact of the model)
\end{itemize}
According to the learning rates, corresponding number of steps was chosen. For instance, if for learning rate = 0.01 100 steps are chosen, then for learning rate = 0.005 twice more steps are needed to converge. The final decision was also seriously influenced by the limits, set on the cHARISMA server\citep{Kostenetskiy_2021}, i.e. 4000 GPU hours per student. Having conducted the experiments, changing learning rates and optimizers of the model, we got that due to the trade-off between the validation error, running time, memory usage and electricity usage, the best optimiser stood out to be SGD with the learning rate = 0.01. Since for each learning rate we got practically the same validation errors with all optimizers, the key benchmarks for the choice were memory and electricity usage, keeping in mind the relation between running time and cHARISMA limits. In addition, lower electricity usage results in economics benefits, saving financial resources as well. For instance, reducing the training by one kilowatt-hour (kWh) saves 0.1\$ in Russia. The detailed results of the research are stored in the Table 2.1.
\begin{table}[h!] 
	\centering 
	\footnotesize	
 \caption[Comparison of Optimisers]
    {\raggedleft{Comparison of Optimisers\par Source: author's calculations}}
		\begin{tabular}{|c|c|c|c|c|}
			\hline
			\textbf{Optimizer} & \textbf{Validation error} & \textbf{Run time } & \textbf{RAM usage } &  \textbf{Electricity usage } \\ 
			 &  & \textbf{(sec)} & \textbf{(MB)} &  \textbf{(kWh)} \\ \hline
			Steps = 100 & ~ & ~ & ~ & ~ \\ \hline
			Learning rate = 0.01 & ~ & ~ & ~ & ~ \\ \hline
			SGD & 3,42 & 181,18 & \underline{\textbf{2627,83}} & \underline{\textbf{0,0214}} \\ \hline
			Adam & 3,42 & 179,03 & 2634,80 & 0,0287 \\ \hline
			RMSProp & 3,42 & \underline{\textbf{177,38}} & 2634,80 & 0,0360 \\ \hline
			Adagrad & \underline{\textbf{3,41}} & 179,79 & 2637,13 & 0,0433 \\ \hline
			~ & ~ & ~ & ~ & ~ \\ \hline
			Steps = 200 & ~ & ~ & ~ & ~ \\ \hline
			Learning rate = 0.01 & ~ & ~ & ~ & ~ \\ \hline
			SGD & 3,30 & 305,74 & \underline{\textbf{2601,52}} & \underline{\textbf{0,0275}} \\ \hline
			Adam & \underline{\textbf{3,28}} & 301,09 & 2606,62 & 0,0338 \\ \hline
			RMSProp & \underline{\textbf{3,28}} & \underline{\textbf{300,78}} & 2606,62 & 0,0420 \\ \hline
			Adagrad & \underline{\textbf{3,28}} & 301,56 & 2605,12 & 0,0527 \\ \hline
			~ & ~ & ~ & ~ & ~ \\ \hline
			Steps = 200 & ~ & ~ & ~ & ~ \\ \hline
			Learning rate = 0.005 & ~ & ~ & ~ & ~ \\ \hline
			SGD & 3,33 & 321,30 & 2668,25 & \underline{\textbf{0,0313}} \\ \hline
			Adam & 3,33 & 312,20 & 2768,57 & 0,0413 \\ \hline
			RMSProp & 3,33 & 310,47 & 2672,90 & 0,0500 \\ \hline
			Adagrad & \underline{\textbf{3,32}} & \underline{\textbf{306,99}} & \underline{\textbf{2588,65}} & 0,0612 \\ \hline
			~ & ~ & ~ & ~ & ~ \\ \hline
			Steps = 2000 & ~ & ~ & ~ & ~ \\ \hline
			Learning rate = 0.0005 &~ & ~ & ~ & ~ \\ \hline
			SGD & \underline{\textbf{3,22}} & 2712,93 & 2764,11 & \underline{\textbf{0,0855}} \\ \hline
			Adam & \underline{\textbf{3,22}} & 2666,55 & 2770,58 & 0,1719 \\ \hline
			RMSProp & \underline{\textbf{3,22}} & \underline{\textbf{2645,44}} & \underline{\textbf{2271,84}} & 0,2604 \\ \hline
			Adagrad & \underline{\textbf{3,22}} & 2674,46 & 2290,94 & 0,3482 \\ \hline
		\end{tabular}
	\label{optimizers_res}
\end{table}

\section{Full Model Fine-Tuning}
For the final GPT-2 fine-tuning we employed the parameters picked previously in the Section \ref{evaluation}, i.e. learning\_rate = 0.01 and SGD as an optimizer. Also the full text corpus of 2.5 GBs was divided into training and test parts.

I launched the training process with 4 GPUs and the total training time was 400 hours (16.5 days). The code of the model training is presented in the Appendix 4.4. The final model was named MS-GPT. 

\subsection{Evaluation of the Full-Trained Model}
To compare the final fine-tuned model with the initial GPT-2, we employed several metrics, used in the text generation validation:
\begin{itemize}
    \item BLEU (Bilingual Evaluation Understudy)
    
    BLEU scores are widely used to evaluate the quality of machine-generated information, especially in the case of machine translation. It measures the similarity between the generated response and one or more reference responses. BLEU’s formula is explained as follows:
\begin{equation}
\text{BLEU} = \text{BP} \times \exp\left(\sum_{n=1}^{N} w_n \cdot \log(\text{precision}_n)\right)
\end{equation}, where:
\begin{itemize}
\item (\text{BP}) is the brevity penalty to account for short generated responses, $BP = min(1, \frac{output length}{reference length})$.
\item (N) is the maximum n-gram order considered.
\item (w\_n) is the weight assigned to the precision of n-grams.
\item (\text{precision}\_n) is the precision of n-grams.
\end{itemize}
To illustrate, let's consider a toy example for BLEU computation. Let's assume that for unigrams, bigrams, and trigrams, the precision is 0.8, 0.6, and 0.4 respectively, and the weights assigned to these precisions are all 1/3. To compute unigram precision, we count the number of individual words that are present in both the generated response and the reference response.We then divide this count by the total number of words in the generated response The same logic holds for bigrams and trigrams also. Let's also assume the brevity penalty (BP) is 1. 

Then, substituting these values into the BLEU formula:

\begin{equation}
    \text{BLEU} = 1 \times \exp\left(\frac{1}{3} \cdot \log(0.8) + \frac{1}{3} \cdot \log(0.6) + \frac{1}{3} \cdot \log(0.4)\right)
\end{equation}

\begin{equation}
\text{BLEU} = 1 \times \exp\left(\frac{1}{3} \cdot (-0.223) + \frac{1}{3} \cdot (-0.511) + \frac{1}{3} \cdot (-0.916)\right)
\end{equation}

\begin{equation}
\text{BLEU} = 1 \times \exp(-0.55)
\end{equation}

\begin{equation}
\text{BLEU} \approx 0.576
\end{equation}

So, in this example, the BLEU score for the generated translation would be approximately 0.576.

\item ROUGE is a set of metrics for evaluating the quality of data collected or generated. This includes measures such as ROUGE-N (n-gram overlap) and ROUGE-L (longest repetitive subsequence). The ROUGE-N metric is defined as:

\begin{equation}
\text{ROUGE-N} = \frac{\sum_{\text{reference}}\sum_{n\text{-grams}} \text{Count}{\text{match}}}{\sum{\text{reference}}\sum_{n\text{-grams}} \text{Count}_{\text{reference}}}
\end{equation}, where:
\begin{itemize}
\item (\text{Count}{\text{match}}) is the count of matching n-grams in the generated response and reference.
\item (\text{Count}{\text{reference}}) is the count of n-grams in the reference.
\end{itemize}
Let's continue with an example for ROUGE-N:

Consider the same example sentence "The cat is on the mat." and its reference translation "The cat is sitting on the mat." We want to compute the ROUGE-1 score for this translation.

First, we count the number of unigrams (individual words) that appear in both the generated response and the reference:

"The": Appears in both.

"cat": Appears in both.

"is": Appears in both.

"on": Appears in both.

"the": Appears in both.

"mat": Appears in both.

So, there are 6 matching unigrams. The total number of unigrams in the reference is 7.

Therefore, the ROUGE-1 score is:

\begin{equation}
    \text{ROUGE-1} = \frac{6}{7} \approx 0.857
\end{equation}

This indicates that 85.7\% of the unigrams in the generated response match with those in the reference.

\item Perplexity is a metric commonly used in language modelling to evaluate how well a probability distribution predicts a sample. For a language model, perplexity is calculated as:

\begin{equation}
\text{Perplexity} = 2^{-\frac{1}{N}\sum_{i=1}^{N} \log_2 P(w_i)}
\end{equation}, where:
\begin{itemize}
\item (N) is the number of words in the generated response.
\item (P(w\_i)) is the probability assigned to the word (w\_i) by the language model.
\end{itemize}
Continuing with our example, let's say we have a language model that assigns the following probabilities to each word in the generated response:

(P(\text{"The"}) = 0.8)

(P(\text{"cat"}) = 0.7)

(P(\text{"is"}) = 0.6)

(P(\text{"on"}) = 0.5)

(P(\text{"the"}) = 0.8)

(P(\text{"mat"}) = 0.9)
Substituting these values into the perplexity formula:

\begin{equation}
    \text{Perplexity} = 2^{-\frac{1}{6}(\log_2(0.8) + \log_2(0.7) + \log_2(0.6) + \log_2(0.5) + \log_2(0.8) + \log_2(0.9))}
\end{equation}

\begin{equation}
    \text{Perplexity} = 2^{-\frac{1}{6}(-0.322 - 0.514 - 0.737 - 1.0 - 0.322 - 0.152)}
\end{equation}

\begin{equation}
    \text{Perplexity} = 2^{-0.509}
\end{equation}

\begin{equation}
    \text{Perplexity} \approx 1.28
\end{equation}

So, the perplexity of this language model for the given response would be approximately 1.28.
\end{itemize}

These examples illustrate how each metric works and how they can be applied to evaluate the quality of machine-generated responses.
These measures provide valuable insights into various aspects of model performance, including fluency, coherence, and similarity in contextual responses The combination of these metrics can provide detailed analysis when comparing with GPT-LLMs. More specifically, BLEU and ROUGE measure response similarity, whereas perplexity measures the ability of the model to predict a given response distribution.

To measure and compare the models' (GPT-2 and MS-GPT) performance we required a certain amount of reference sentences to compute the aforepresented metrics between the generated lines and their respective sources. For this purpose, we needed to retrieve some ground truth in the field of mass spectrometry, not any private opinion, but clear facts, which can be established using more or less the same set of words and phrases. That is why we took a bunch of examples of random definitions from the mass spectrometry dictionary\citep{msdict}. Definitions, used for models' evaluation are presented below:
\begin{itemize}
    \item Bottom-up proteomics is a method of protein identification that uses proteolytic digestion before analysis by liquid chromatography and mass spectrometry.
    \item Imaging mass spectrometry is a procedure used to form chemically selective images of an object based on the mass spectrometric detection of ions desorbed from its surface.
    \item Autodetachment is a process whereby a negative ion in a discrete state with energy greater than the detachment threshold loses an electron spontaneously without further interaction with an energy source.
    \item Gas chromatography-mass spectrometry (GC-MS) is a technique by which a mixture is separated into individual components by gas chromatography, followed by detection with a mass spectrometer.
    \item Mass spectrometry is study of matter through the formation of gas-phase ions that are characterized using mass spectrometers by their mass, charge, structure, and/or physico-chemical properties.
    \item Peptide mass fingerprinting (PMF) is a method for protein analysis where an unknown protein is chemically or enzymatically cleaved into peptide fragments whose masses are determined by mass spectrometry.
    \item Peptide sequence tag is sequence of a peptide ion fragment masses that can be used to aid in the identification of the amino acid sequence.
    \item Photodissociation is a process wherein the reactant ion or molecule is dissociated as a result of absorption of one or more photons.
    \item Pneumatically assisted electrospray ionization is electrospray ionization in which the nebulization of the liquid stream is assisted by a concentric stream of gas.
    \item Post-acceleration detector (PAD) is a detector in which a high potential is applied after m/z separation to accelerate the ions and produce an improved signal.
    \item Product ion spectrum is a mass spectrum in which the appropriate m/z separation analysis function is set to record the product ions of a selected precursor ion selected by m/z".
    \item Collision cell is a chamber in the ion path between m/z separation elements, or between ion source acceleration region and the first analyzer, in tandem mass spectrometry in space configurations.
    \item Collision quadrupole is a transmission quadrupole to which an oscillating radio frequency potential is applied so as to focus a beam of ions through a collision gas or buffer gas with no m/z separation other than low m/z cut-off.
    \item Continuous-flow fast atom bombardment (CF-FAB) is a variant of fast atom bombardment in which the mixture of analyte and liquid matrix is continuously supplied to the sample probe tip.
    \item Deconvoluted mass spectrum is a mass spectrum processed with an algorithm designed to extract a desired signal or signals from raw experimental data in which the desired signals have been complicated (convolved) by some interferences or in some other way.
    \item Direct infusion is a method of liquid sample introduction in which the sample is continuously flowed into a mass spectrometer ion source.
    \item Dynamic exclusion is a software method used to minimize repeat selections of identical precursor ions for collision-induced dissociation in replicate chromatography-tandem mass spectrometry analyses of complex mixtures.
    \item Hydrogen/deuterium exchange (HDX) is an exchange of hydrogen atoms with deuterium atoms in a chemical species in solution prior to introduction into a mass spectrometer, or by ion/molecule reaction with a neutral gas inside a mass spectrometer.
    \item Hyphenated mass spectrometry technique is an analytical technique in which mass spectrometry is interfaced with a pretreatment step, most often chromatographic separation but many other combinations are possible.
    \item Inductively coupled plasma-mass spectrometry (ICP-MS) is a mass spectrometry technique based on coupling a mass spectrometer with an inductively coupled plasma as an ion source that both atomizes samples into their constituent atoms and ionizes them to form atomic cations.
    \item Ion funnel is series of stacked ring electrodes with progressively decreasing inner diameter to which a combined radio frequency and fixed potential is applied.
    \item Ion gate is a set of plates or grid of wires in an ion mobility spectrometer, time-of-flight mass spectrometer, or other mass spectrometer that is used to apply a pulsed electric field with the purpose of selectively deflecting charged particles.
    \item Ionic dissociation is a dissociation of an ion into another ion of lower mass and one or more neutral or ions with a lower charge.
    \item Charge number, z is an absolute value of charge of an ion divided by the value of the elementary charge (e) rounded to the nearest integer.
    \item Static secondary ion mass spectrometry (SSIMS) is a method of secondary ion mass spectrometry using low current densities for analysis of sample surface components, in contrast with dynamic secondary ion mass spectrometry.
    \item Tandem mass spectrometer is a mass spectrometer designed for mass spectrometry.
    \item Accelerating potential is an electrical potential difference used to impart translational energy to ions.
    \item Accelerator mass spectrometry (AMS) is a mass spectrometry technique in which atoms and molecules from a sample are ionized, accelerated to MeV energies and separated according to their momentum, charge, and energy, allowing high discrimination for measurement of isotope abundances.
    \item Auxiliary gas is a gas used in a spray ion source in addition to the nebulizing gas to aid in solvent removal.
    \item Beam mass spectrometer is a mass spectrometer in which an ion beam accelerated from the ion source is transmitted through a m/z analyzer, or analyzers, to the detector.
\end{itemize}
The total number of instances is 30. Since we are dealing with GPT models, we need to pass some special prompt in order to give some instructions for the model to complete the task, e.g. for the definitions above, the prompts, passed to the models were the following:
\begin{itemize}
    \item Give the definition: Bottom-up proteomics is
    \item Give the definition: Data-dependent acquisition is
    \item Give the definition: Imaging mass spectrometry is
    \item etc.
\end{itemize}
The detailed code is presented in the Appendix 4.7.

\subsection{Statistical Testing}
After passing all prompts through the competing models, we computed the following scores: BLEU, ROUGE-1, ROUGE-L and Perplexity. The detailed code is presented in Appendix 4.6. In order to find out whether the differences between models are statistically significant, we conducted Wilcoxon rank-sum statistical test\citep{wilcoxon} due to the paired data with the following hypotheses:

BLEU:
\begin{itemize}
    \item $H_{0}: BLEU_{MS-GPT} = BLEU_{GPT-2}$
    \item $H_{1}: BLEU_{MS-GPT} > BLEU_{GPT-2}$
\end{itemize}
ROUGE-1:
\begin{itemize}
    \item $H_{0}: ROUGE-1_{MS-GPT} = ROUGE-1_{GPT-2}$
    \item $H_{1}: ROUGE-1_{MS-GPT} > ROUGE-1_{GPT-2}$
\end{itemize}
ROUGE-L:
\begin{itemize}
    \item $H_{0}: ROUGE-L_{MS-GPT} = ROUGE-L_{GPT-2}$
    \item $H_{1}: ROUGE-L_{MS-GPT} > ROUGE-L_{GPT-2}$
\end{itemize}
 Perplexity:
\begin{itemize}
    \item $H_{0}: Perplexity_{MS-GPT} = Perplexity_{GPT-2}$
    \item $H_{1}: Perplexity_{MS-GPT} < Perplexity_{GPT-2}$
\end{itemize}
The tests are one-sided and significance level is 5\%. We reject $H_{0}$ and accept the alternative if p-value is less than 0.05. According to the average values, presented in the Table \ref{tab:metrics}, MS-GPT outperforms GPT-2 across all of the metrics, we can conclude that statistically significant it is better only in terms of Perplexity, i.e. MS-GPT generates more coherent text.

\begin{table}[h!]
\begin{center}
    \caption[Average scores for GPT-2 and MS-GPT]
    {\raggedleft{Average scores for GPT-2 and MS-GPT\par Source: author's calculations}}
    \label{tab:metrics}
    \begin{tabular}{|c|c|c|c|c|} 
    \hline
      \textbf{Model} & \textbf{BLEU Score} & \textbf{ROUGE-1 Score} & \textbf{ROUGE-L Score} & \textbf{Perplexity Score} \\
      \hline
      GPT-2 & 0.33 & 0.42 & 0.57 & 13586.37 \\
      \hline
      MassSpecGPT & \underline{\textbf{0.34}} & \underline{\textbf{0.44}} & \underline{\textbf{0.62}} & \underline{\textbf{10092.12}} \\
      \hline
    \end{tabular}
\end{center}
\end{table}

The results of Wilcoxon rank-sum test calculations for metrics BLEU, ROUGE-1, ROUGE-L, Perplexity are presented in the Table \ref{tab:test}.

\begin{table}[h!]
\begin{center}
    \caption[Wilcoxon rank-sum test results]{\raggedleft{Wilcoxon rank-sum test results\par Source: author's calculations}}  
    \label{tab:test}
    \begin{tabular}{|c|c|c|} 
    \hline
      \textbf{Metric} & \textbf{Test statistic} & \textbf{P-value} \\
      \hline
      BLEU & 244.0 & 0.60 \\
      \hline
      ROUGE-1 & 191.0 & 0.88 \\
      \hline
      ROUGE-L & 196.5 & 0.82 \\
      \hline
      Perplexity & 106.0 & \underline{\textbf{0.004}} \\
      \hline
    \end{tabular}
\end{center}
\end{table}

Another point of study is to measure and compare the models' reproducibilities. For this purpose, we embedded the phrase "in mass spectrometry" in every prompt where it makes sense.

\begin{itemize}
    \item Give the definition: Bottom-up proteomics in mass spectrometry is
    \item Give the definition: Data-dependent acquisition in mass spectrometry is
    \item Give the definition: Imaging mass spectrometry is
    \item Give the definition: Ion funnel in mass spectrometry is
    \item etc. 
\end{itemize}
Thereafter, we launched the same code as in the Appendix 4.7 and wrote down the resulting sentences in the separate files for each of the models. 

Afterwards, we employed BERT model\citep{devlin2019bert} to compute the embeddings of the texts, generated by each model based on the bare prompts and the ones with the phrase "in mass spectrometry" modification. The reproducibility score was interpreted and computed as cosine similarity between the respective embeddings\citep{Bingyu_2022}. In essence, the process of cosine similarity computations is based on the measuring the cosine of the angle between two vectors, i.e. sentences embeddings. Hence, the score lies between -1 and +1, where -1 means that the sentences are opposite in their sense whereas +1 depicts reverse case.
The code of these computations is presented in the Appendix 4.5.
The result of the reproducibility measurements are presented in the Table \ref{tab:reproducibility_metrics}.

\begin{table}[h!]
\begin{center}
    \caption[Reproducibility scores for GPT-2 and MS-GPT]{\raggedleft{Reproducibility scores for GPT-2 and MS-GPT\par Source: author's calculations}}  
    \label{tab:reproducibility_metrics}
    \begin{tabular}{|c|c|} 
    \hline
      \textbf{Model} & \textbf{Reproducibility Score} \\
      \hline
      GPT-2 & 0.83 \\
      \hline
      MassSpecGPT & \underline{\textbf{0.84}} \\
      \hline
    \end{tabular}
\end{center}
\end{table}
\newpage

Afterwards, we conducted the computations of the Wilcoxon statistics to find out the significance of the difference between the models, the code is presented in the Appendix 4.8. The resulting p-value was 0.03, which means that there are statistically significant improvements in the GPT-2 behaviour under the 5\% level.

\chapter{Summary and Conclusion}

\section{Accuracy Improvements}
Although the final model MS-GPT has shown a solid result comparing to the original GPT-2, still there are paths that can improve the accuracy and coherence of the model, fine-tuned for the field of mass spectrometry.

For instance, there could be conducted a human-base supervision of the training text corpus. In our research we used some elimination of the unnecessary parts of the text, e.g. such as bibliographies, unreadable formulas, etc. By the detailed investigation of the text, the corpus would be more clean and natural as the training base.

Another point for the enhancement is to pick another GPT-like model as the base one for the further fine-tuning. On the HuggingFace there are a few models with the same objectives as GPT-2 has, by measuring and comparing the initial metrics of each, the best one should be chosen.

Thirdly, apart from applying the LoRA adapters, some additional layers could be constructed on the top of existing architectures. This point depends on the data volume, available for the training and the computational power that a researcher disposes.

\section{Summary of Contributions}
In the current research we have presented the development of a novel text generating model, named MS-GPT in the fields of mass spectrometry, being the very first model, fine-tuned for this narrow scope. During the research, we used various methods and techniques that lead us to the final result:
\begin{itemize}
    \item Semantic API key for the paper bulk downloading;
    \item Containerization application "Docker" to launch the ML-based approach of text converting from PDF format;
    \item The Linux servers to operate the files and parallelism of the PDF preprocessing process;
    \item LoRA adapters to fine-tune the OpenAI GPT-2 model with our text corpus, preliminary divided by training and test parts;
    \item HSE cHARISMa supercomputer, equipped by several GPUs to train MS-GPT;
    \item Novel method of comparing GPT-like large language models' results by computing the non-parametric statistical criteria Wilcoxon sum-ranked test
    \item Novel technique of a model's reproducibility computation based on the cosine similarity between the outputs
\end{itemize}
As the result, MS-GPT outperforms GPT-2 in Perplexity score, which measures the coherence of the model. In addition, we achieved statistically significant improvement of the reproducibility under 5\% significance level.

\section{Future Research Directions}
In the research, we developed a decoder-only model, targeting the text generation based on the prompt. 
The proposed methods and techniques open up a way for the development of MS-Chat-GPT, which consists of an encoder part alongside with a decoder module. This kind of models is related to the conversational types and can hold up a dialogue with the user. The production of a such model requires large text corpus and the extraction of the question-answer (QA) pairs to build a dataset the further would be used in the training process.
The base for such a model could also be used from the Hugging Face space.

Besides, the developed MS-GPT model could be used by the mass spectrometry labs to complete scientific texts and help students with their writings/experiments by providing more plausible information.

\chapter{Appendices}

\raggedleft{\section{Appendix A: Papers Downloading Code}}
\label{appendix:paper_downloading}

\begin{lstlisting}[language=Python, caption=Python example,
label=lst:python_downloading,
basicstyle=\ttfamily\small,
keywordstyle=\bfseries\color{blue},
commentstyle=\itshape\color{gray},
stringstyle=\color{red},
numbers=left, numberstyle=\tiny\color{gray},
breaklines=true, breakatwhitespace=true,
frame=single,
backgroundcolor=\color{white}]
import json
import requests
import os
import time
import multiprocessing 
file_path = "papers_total.jsonl"
output_folder = "/blob/dasulimov/"
resume_index = 0  
def download_pdf(url, file_path):
    response = requests.get(url, verify=False)
    response.raise_for_status()
    with open(file_path, "wb") as file:
        file.write(response.content)
def resume_download(url, file_path, offset):
    resume_header = {"Range": f"bytes={offset}-"}
    response = requests.get(url, headers=resume_header,
    stream=True, verify=False, allow_redirects=True)
    with open(file_path, "ab") as file:
        file.write(response.content)
def process_line(input_tuple):
    line, index = input_tuple
    data = json.loads(line)
    pdf_info = data.get("openAccessPdf")
    if pdf_info and pdf_info.get("url"):
        paper_id = data.get("paperId")
        pdf_url = pdf_info["url"]
        pdf_file_path = f"{output_folder}{paper_id}.pdf"
        offset = 0  
        try:
            if os.path.exists(pdf_file_path):
                offset = os.path.getsize(pdf_file_path) 
                resume_download(pdf_url, pdf_file_path, offset)  
            else:
                download_pdf(pdf_url, pdf_file_path)
            print(f"PDF downloaded for paper ID: {paper_id}")
        except requests.exceptions.RequestException as e:
            print(f"Error downloading PDF for paper ID: {paper_id}")
            print(f"Error message: {str(e)}")
    return index
if os.path.exists("/blob/resume.txt"):
    with open("/blob/resume.txt", "r") as resume_file:
        resume_index = int(resume_file.read())
lines = []
with open(file_path, "r") as file:
    for i, line in enumerate(file):
        if i < resume_index:
            continue  
        lines.append((line, i))
pool = multiprocessing.Pool()
for result in pool.imap_unordered(process_line, lines):
    with open("/blob/resume.txt", "w") as resume_file:
        resume_file.write(str(result))

\end{lstlisting}
\newpage

\raggedleft{\section{Appendix B: Text Extraction Code}}
\label{appendix:text_extraction}

\begin{mycode}[caption={Python code for text extraction},label=lst:python,
basicstyle=\ttfamily\small,
keywordstyle=\bfseries\color{blue},
commentstyle=\itshape\color{gray},
stringstyle=\color{red},
numbers=left, numberstyle=\tiny\color{gray},
breaklines=true, breakatwhitespace=true,
frame=single,
backgroundcolor=\color{white}]
import os
import requests
from bs4 import BeautifulSoup
import logging
import time
import chardet

def process_pdf(paper):
    url = 'http://localhost:8070/api/processFulltextDocument'
    multipart = 
    {'input': (os.path.basename(paper), open(paper, 'rb'), 'application/pdf')}
    response = 
    requests.post(url, files=multipart, headers={'Accept': 'application/xml'})
    if response.status_code == 200:
        return response.text
    else:
        print(f'Error processing file {paper}: status code {response.status_code}')
pdf_directory = '/blob/dasulimov'
output_file_path = '/blob/grobid_llm.txt'
count = 0
with open(output_file_path, 'a', encoding='utf-8') as output_file:
    for file_name in os.listdir(pdf_directory):
        logging.info(file_name)
        if file_name.endswith('.pdf'):
            try:
                pdf_path = os.path.join(pdf_directory, file_name)
                pdf_data = process_pdf(pdf_path)
                bs = BeautifulSoup(pdf_data, 'lxml')
                text = ' '.join(p.get_text() for p in bs.find_all('p'))
                encoding = chardet.detect(text.encode())['encoding']
                if encoding == 'ascii' or encoding == 'utf-8':
                    output_file.write(text + '\n')
                    output_file.write('\n')
                    count += 1
                    print(f'Have preprocessed paper number {count}')
                else:
                    print(f'Skipped paper number {count} due to unreadable format')
            except Exception as e:
                print(f"An error occurred: {e}")
                continue

\end{mycode}
\newpage

\raggedleft{\section{Appendix C: Parameter Optimization Code}}
\label{appendix:optimization_parameters}

\begin{mycode}[caption={Python code for choosing the best optimization parameters},label=lst:python,
basicstyle=\ttfamily\small,
keywordstyle=\bfseries\color{blue},
commentstyle=\itshape\color{gray},
stringstyle=\color{red},
numbers=left, numberstyle=\tiny\color{gray},
breaklines=true, breakatwhitespace=true,
frame=single,
backgroundcolor=\color{white},
]
import torch.optim as optim
from transformers import (GPT2Tokenizer, GPT2LMHeadModel, TextDataset,
                          DataCollatorForLanguageModeling, Trainer,
                          TrainingArguments, EarlyStoppingCallback)
import matplotlib.pyplot as plt
import numpy as np
import os
import torch
from transformers import AutoTokenizer, AutoConfig, AutoModelForCausalLM
from peft import LoraConfig, get_peft_model,  prepare_model_for_int8_training 
import transformers
os.chdir('/home/dasulimov')
model_name = "gpt2"
tokenizer = AutoTokenizer.from_pretrained(model_name)
device = torch.device('cuda' if torch.cuda.is_available() else 'cpu')
model = AutoModelForCausalLM.from_pretrained(
    model_name
).to(device)
config = LoraConfig(
    r=4,
    lora_dropout=0.05,
    bias="none",
    target_modules=["c_attn", "c_proj", "lm_head"],
    task_type="CAUSAL_LM"
)
for param in model.parameters():
    param.requires_grad = False  
model.gradient_checkpointing_enable()  
model.enable_input_require_grads()
peft_model = get_peft_model(model, config)
train_parameters = 0  
total_params = 0 
for _, param in peft_model.named_parameters():
    total_params += param.numel()  
    if param.requires_grad:  
        train_parameters += param.numel()  
print(f"Trainable params: {train_parameters}")  
print(f"All params: {total_params}")  
print(f"Trainable: {100 * train_parameters / total_params:.2f}\%")  
train_dataset = TextDataset(
    tokenizer=tokenizer,
    file_path='grobid_demo.txt',
    block_size=128,
)
eval_dataset = TextDataset(
    tokenizer=tokenizer,
    file_path='grobid_demo.txt',
    block_size=128,
)
data_collator = DataCollatorForLanguageModeling(
    tokenizer=tokenizer,
    mlm=False  # Set mlm to False for text generation
)
lr = [1e-2, 5e-3, 5e-4]
optimizers = [
    optim.SGD(peft_model.parameters(), lr=learning_rate) for learning_rate in lr
] + [
    optim.Adam(peft_model.parameters(), lr=learning_rate) for learning_rate in lr
] + [
    optim.RMSprop(peft_model.parameters(), lr=learning_rate) for learning_rate in lr
] + [
    optim.Adagrad(peft_model.parameters(), lr=learning_rate) for learning_rate in lr
]
best_loss = float('inf')
best_optimizer = None
best_learning_rate = None
best_model_state_dict = None
original_model_state_dict = peft_model.state_dict()
for optimizer in optimizers:
    peft_model.load_state_dict(original_model_state_dict)
    training_args = TrainingArguments(
    output_dir='./output',  
    overwrite_output_dir=True,
    max_steps=400,  
    per_device_train_batch_size=4,  
    save_total_limit=2,  
    logging_steps=40, 
    evaluation_strategy="steps",
    learning_rate=optimizer.param_groups[0]['lr'],  # Learning rate
    eval_steps=40,  
    logging_dir='./logs', 
    gradient_accumulation_steps=16,  
    weight_decay=0.01,  
    fp16=True,  
    dataloader_num_workers=4  
)
    trainer = Trainer(
        model=peft_model,
        args=training_args,
        data_collator=data_collator,
        train_dataset=train_dataset,
        eval_dataset=eval_dataset
    )
    trainer.train()
    training_loss = []
    test_loss = []
    for el in trainer.state.log_history:
        if 'loss' in el:
            training_loss.append(el['loss'])
        elif 'eval_loss' in el:
            test_loss.append(el['eval_loss'])
    steps = np.arange(0, len(training_loss) * trainer.state.logging_steps, 
    trainer.state.logging_steps)
    plt.title('Loss Plot for {}'.format(optimizer.__class__.__name__))
    plt.plot(steps, training_loss, label='Training Loss')
    plt.plot(steps, test_loss, label='Evaluation Loss')
    plt.xlabel('Steps')
    plt.ylabel('Loss')
    plt.legend()
    plt.grid()
    plt.savefig('./plots/loss_plot_{}_{}.png'.\
    format(optimizer.__class__.__name__, 
    optimizer.param_groups[0]["lr"]))
    print("Evaluation loss with {} and learning rate {}: {}".
    format(optimizer.__class__.__name__, 
    optimizer.param_groups[0]['lr'], test_loss[-1]))
    if test_loss[-1] < best_loss:
        best_loss = test_loss[-1]
        best_optimizer = optimizer
        best_learning_rate = optimizer.param_groups[0]['lr']
        best_model_state_dict = peft_model.state_dict()
peft_model.load_state_dict(best_model_state_dict)
peft_model.transformer.wte.weight = 
torch.nn.Parameter(peft_model.transformer.\ 
wte.weight.clone().detach())
peft_model.lm_head.base_layer.weight = 
torch.nn.Parameter(peft_model.lm_head.base_layer.\ weight.clone().detach())
best_trainer = Trainer(
    model=peft_model,
    args=training_args,
    data_collator=data_collator,
    train_dataset=train_dataset,
    eval_dataset=eval_dataset
)
best_trainer.save_model('./models/best_model_{}_{}.pt'.
format(best_optimizer.__class__.__name__, best_learning_rate))
tokenizer.save_pretrained('./models/best_model_{}_{}.pt'.
format(best_optimizer.__class__.__name__, best_learning_rate))
print("Best model achieved an evaluation loss of {} 
with {} optimizer and a learning rate of {}.".
format(best_loss, best_optimizer.__class__.__name__, best_learning_rate))

\end{mycode}
\newpage
\raggedleft{\section{Appendix D: GPT-2 Fine-Tuning Code}}
\label{appendix:fine_tuning}

\begin{mycode}[caption={Python code for final GPT-2 fine-tuning},label=lst:python,
basicstyle=\ttfamily\small,
keywordstyle=\bfseries\color{blue},
commentstyle=\itshape\color{gray},
stringstyle=\color{red},
numbers=left, numberstyle=\tiny\color{gray},
breaklines=true, breakatwhitespace=true,
frame=single,
backgroundcolor=\color{white},
]
import torch.optim as optim
from transformers import (GPT2Tokenizer, GPT2LMHeadModel, TextDataset,
                          DataCollatorForLanguageModeling, Trainer,
                          TrainingArguments)
import numpy as np
import os
import torch
from transformers import AutoTokenizer, AutoModelForCausalLM
from peft import LoraConfig, get_peft_model
from tqdm.notebook import tqdm
# Set your working directory
os.chdir('/home/dasulimov')
model_name = "gpt2"
tokenizer = GPT2Tokenizer.from_pretrained(model_name, max_length=2048)
device = torch.device('cuda' if torch.cuda.is_available() else 'cpu')
model = AutoModelForCausalLM.from_pretrained(
    model_name
).to(device)
config = LoraConfig(
    r=4,
    lora_dropout=0.05,
    bias="none",
    target_modules=["c_attn", "c_proj", "lm_head"],
    task_type="CAUSAL_LM"
)
peft_model = get_peft_model(model, config)
train_parameters = 0 
total_params = 0  
for _, param in peft_model.named_parameters():
    total_params += param.numel()  
    if param.requires_grad:  
        train_parameters += param.numel()  
train_dataset = TextDataset(
    tokenizer=tokenizer,
    file_path='grobid_llm.txt',
    block_size=128,
)
eval_dataset = TextDataset(
    tokenizer=tokenizer,
    file_path='grobid_llm.txt',
    block_size=128,
)
data_collator = DataCollatorForLanguageModeling(
    tokenizer=tokenizer,
    mlm=False  
)
num_steps = 100000
learning_rate = 0.01
training_args = TrainingArguments(
    output_dir='/home/dasulimov/output',  
    overwrite_output_dir=True,
    max_steps=num_steps,  
    per_device_train_batch_size=4,  
    save_total_limit=2,  
    logging_steps=10000,  
    evaluation_strategy="steps",
    learning_rate=learning_rate, 
    eval_steps=10000,  
    logging_dir='/home/dasulimov/logs', 
    gradient_accumulation_steps=16, 
    weight_decay=0.01,  
    fp16=True,  
    dataloader_num_workers=4,
    save_steps=10000
)
optimizer = optim.SGD(peft_model.parameters(), lr=learning_rate) 
trainer = Trainer(
    model=peft_model,
    args=training_args,
    data_collator=data_collator,
    train_dataset=train_dataset,
    eval_dataset=eval_dataset,
    optimizers=(optimizer, None)
)
trainer.train()

\end{mycode}
\newpage

\raggedleft{\section{Appendix E: Reproducibility Calculation}}
\label{appendix:reproducibility}

\begin{mycode}[caption={Python code for model's reproducibility calculation},label=lst:python,
basicstyle=\ttfamily\small,
keywordstyle=\bfseries\color{blue},
commentstyle=\itshape\color{gray},
stringstyle=\color{red},
numbers=left, numberstyle=\tiny\color{gray},
breaklines=true, breakatwhitespace=true,
frame=single,
backgroundcolor=\color{white},
]
import torch
from transformers import BertModel, BertTokenizer
from sklearn.metrics.pairwise import cosine_similarity
model_name_or_path = 'bert-base-uncased'
tokenizer = BertTokenizer.from_pretrained(model_name_or_path)
model = BertModel.from_pretrained(model_name_or_path)
def compute_bert_embeddings(texts):
    inputs = tokenizer(texts, padding=True, truncation=True, return_tensors="pt")
    with torch.no_grad():
        outputs = model(**inputs)
        embeddings = outputs.last_hidden_state[:, 0, :]  # Extract the [CLS] token embeddings
    return embeddings.numpy()
model_names = ["masspecgpt", "gpt2"]
original_embeddings = {}
for model_name in model_names:
    with open(f"{model_name}_generated_text.txt", "r") as file:
        original_texts = file.readlines()
        original_embeddings[model_name] = compute_bert_embeddings(original_texts)
synonym_embeddings = {}
for model_name in model_names:
    with open(f"{model_name}_generated_text_synonyms.txt", "r") as file:
        synonym_texts = file.readlines()
        synonym_embeddings[model_name] = compute_bert_embeddings(synonym_texts)
for model_name in model_names:
    similarities = []
    for i in range(len(original_embeddings[model_name])):
        cosine_similarities = cosine_similarity(original_embeddings[model_name]\
        [i].reshape(1, -1), synonym_embeddings[model_name][i].reshape(1, -1))
        similarities.append(cosine_similarities)
    with open(f"{model_name}_reproducibility_scores.csv", 'w') as file:
        file.write("Pair Number,Cosine Similarity\n")
        for i, cosine_similarity in enumerate(similarities):
            file.write(f"{i+1},{cosine_similarity[0][0]}\n")


\end{mycode}
\newpage

\raggedleft{\section{Appendix F: Statistics Calculation of Main Models' Metrics}}
\label{appendix:statistics}

\begin{mycode}[caption={Python code for statistics calculation of main models' metrics},label=lst:python,
basicstyle=\ttfamily\small,
keywordstyle=\bfseries\color{blue},
commentstyle=\itshape\color{gray},
stringstyle=\color{red},
numbers=left, numberstyle=\tiny\color{gray},
breaklines=true, breakatwhitespace=true,
frame=single,
backgroundcolor=\color{white},
]
import pandas as pd
from scipy.stats import mannwhitneyu
from scipy.stats import wilcoxon
df = pd.read_csv("model_comparison\
_metrics_synonyms_individual.csv")
df_masspecgpt = df[df['Model'] == 'masspecgpt']
df_gpt2 = df[df['Model'] == 'gpt2']
for metric in ['BLEU Score', 'ROUGE-1 Score', 'ROUGE-L Score', 'Perplexity Score']:
    result = mannwhitneyu(df_masspecgpt[metric], df_gpt2[metric], alternative='less')
    print(f"Mann-Whitney U test for {metric}:")
    print(f"Test statistic: {result.statistic}")
    print(f"p-value: {result.pvalue}")
    print()
for metric in ['BLEU Score', 'ROUGE-1 Score', 'ROUGE-L Score', 'Perplexity Score']:
    result = mannwhitneyu(df_masspecgpt[metric], df_gpt2[metric], alternative='less')
    print(f"Mann-Whitney U test for {metric}:")
    print(f"Test statistic: {result.statistic}")
    print(f"p-value: {result.pvalue}")
    print()
for metric in ['BLEU Score', 'ROUGE-1 Score', 'ROUGE-L Score', 'Perplexity Score']:
    result = wilcoxon(df_masspecgpt[metric], df_gpt2[metric], alternative='less')
    print(f"Wilcoxon signed-rank test for {metric}:")
    print(f"Test statistic: {result.statistic}")
    print(f"p-value: {result.pvalue}")
    print()

\end{mycode}
\newpage

\raggedleft{\section{Appendix G: Scientific Text Generation}}
\label{appendix:text_generation}

\begin{mycode}[caption={Python code for scientific text generation},label=lst:python,
basicstyle=\ttfamily\small,
keywordstyle=\bfseries\color{blue},
commentstyle=\itshape\color{gray},
stringstyle=\color{red},
numbers=left, numberstyle=\tiny\color{gray},
breaklines=true, breakatwhitespace=true,
frame=single,
backgroundcolor=\color{white},
]
import torch
from transformers import AutoModelForCausalLM, AutoTokenizer
from torch.nn.utils.rnn import pad_sequence
from nltk.translate.bleu_score import sentence_bleu
import math
from transformers import GPT2Model, GPT2Tokenizer
import pandas as pd
import numpy as np
model_names = ["gpt2", "masspecgpt"]
models = []
tokenizers = []
for model_name in model_names:
    if model_name == "masspecgpt":
        tokenizer = AutoTokenizer.\
        from_pretrained('/home/dasulimov/content/final-tuning-new/fine-tuned-model')
        model = AutoModelForCausalLM.\
        from_pretrained('/home/dasulimov/content/final-tuning-new/fine-tuned-model')
    else:
        tokenizer = AutoTokenizer.from_pretrained(model_name)
        model = AutoModelForCausalLM.from_pretrained(model_name)
    models.append(model)
    tokenizers.append(tokenizer)
# Define multiple prompts and their corresponding target prompts
prompts = [
  ("Give the definition: Autodetachment is",
   "Give the definition: Autodetachment is process whereby a negative ion in a discrete state with energy greater than the detachment threshold loses an electron spontaneously without further interaction with an energy source."),
  ("Give the definition: Bottom-up proteomics is",
   "Give the definition: Bottom-up proteomics is method of protein identification that uses proteolytic digestion before analysis by liquid chromatography and mass spectrometry."),
  ("Give the definition: Gas chromatography-mass spectrometry (GC-MS) is",
   "Give the definition: Gas chromatography-mass spectrometry (GC-MS) is technique by which a mixture is separated into individual components by gas chromatography, followed by detection with a mass spectrometer."),
  ("Give the definition: Mass spectrometry is",
   "Give the definition: Mass spectrometry is study of matter through the formation of gas-phase ions that are characterized using mass spectrometers by their mass, charge, structure, and/or physico-chemical properties."),
  ("Give the definition: Peptide mass fingerprinting (PMF) is",
   "Give the definition: Peptide mass fingerprinting (PMF) is Method for protein analysis where an unknown protein is chemically or enzymatically cleaved into peptide fragments whose masses are determined by mass spectrometry."),
  ("Give the definition: Peptide sequence tag is",
   "Give the definition: Peptide sequence tag is sequence of peptide ion fragment masses that can be used to aid in the identification of the amino acid sequence."),
  ("Give the definition: photodissociation is",
   "Give the definition: photodissociation is process wherein the reactant ion or molecule is dissociated as a result of absorption of one or more photons."),
  ("Give the definition: pneumatically assisted electrospray ionization is",
   "Give the definition: pneumatically assisted electrospray ionization is electrospray ionization in which the nebulization of the liquid stream is assisted by a concentric stream of gas."),
  ("Give the definition: post-acceleration detector (PAD) is",
   "Give the definition: post-acceleration detector (PAD) is detector in which a high potential is applied after m/z separation to accelerate the ions and produce an improved signal."),   
  ("Give the definition: product ion spectrum is",
   "Give the definition: product ion spectrum is mass spectrum in which the appropriate m/z separation analysis function is set to record the product ions of a selected precursor ion selected by m/z"),
  ("Give the definition: collision cell is",
   "Give the definition: collision cell is chamber in the ion path between m/z separation elements, or between ion source acceleration region and the first analyzer, in tandem mass spectrometry in space configurations."),
  ("Give the definition: collision quadrupole is",
   "Give the definition: collision quadrupole is transmission quadrupole to which an oscillating radio frequency potential is applied so as to focus a beam of ions through a collision gas or buffer gas with no m/z separation other than low m/z cut-off."),
  ("Give the definition: continuous-flow fast atom bombardment (CF-FAB) is",
   "Give the definition: continuous-flow fast atom bombardment (CF-FAB) is variant of fast atom bombardment in which the mixture of analyte and liquid matrix is continuously supplied to the sample probe tip."),
  ("Give the definition: deconvoluted mass spectrum is",
   "Give the definition: deconvoluted mass spectrum is mass spectrum processed with an algorithm designed to extract a desired signal or signals from raw experimental data in which the desired signals have been complicated (convolved) by some interferences or in some other way"),
  ("Give the definition: direct infusion is",
   "Give the definition: direct infusion is method of liquid sample introduction in which the sample is continuously flowed into a mass spectrometer ion source."),
  ("Give the definition: dynamic exclusion is",
   "Give the definition: dynamic exclusion is software method used to minimize repeat selections of identical precursor ions for collision-induced dissociation in replicate chromatography-tandem mass spectrometry analyses of complex mixtures."),
  ("Give the definition: hydrogen/deuterium exchange (HDX) is",
   "Give the definition: hydrogen/deuterium exchange (HDX) is exchange of hydrogen atoms with deuterium atoms in a chemical species in solution prior to introduction into a mass spectrometer, or by ion/molecule reaction with a neutral gas inside a mass spectrometer."),
  ("Give the definition: hyphenated mass spectrometry technique is",
   "Give the definition: hyphenated mass spectrometry technique is analytical technique in which mass spectrometry is interfaced with a pretreatment step, most often chromatographic separation but many other combinations are possible."),
  ("Give the definition: imaging mass spectrometry is",
   "Give the definition: imaging mass spectrometry is procedure used to form chemically selective images of an object based on the mass spectrometric detection of ions desorbed from its surface."),
  ("Give the definition: inductively coupled plasma-mass spectrometry (ICP-MS) is",
   "Give the definition: inductively coupled plasma-mass spectrometry (ICP-MS) is mass spectrometry technique based on coupling a mass spectrometer with an inductively coupled plasma as an ion source that both atomizes samples into their constituent atoms and ionizes them to form atomic cations."),
  ("Give the definition: ion funnel is",
   "Give the definition: ion funnel is series of stacked ring electrodes with progressively decreasing inner diameter to which a combined radio frequency and fixed potential is applied."),
  ("Give the definition: ion gate is",
   "Give the definition: ion gate is set of plates or grid of wires in an ion mobility spectrometer, time-of-flight mass spectrometer, or other mass spectrometer that is used to apply a pulsed electric field with the purpose of selectively deflecting charged particles."),
  ("Give the definition: ionic dissociation is",
   "Give the definition: ionic dissociation is dissociation of an ion into another ion of lower mass and one or more neutral or ions with a lower charge."),
  ("Give the definition: charge number, z is",
   "Give the definition: charge number, z is absolute value of charge of an ion divided by the value of the elementary charge (e) rounded to the nearest integer."),
  ("Give the definition: static secondary ion mass spectrometry (SSIMS) is",
   "Give the definition: static secondary ion mass spectrometry (SSIMS) is method of secondary ion mass spectrometry using low current densities for analysis of sample surface components, in contrast with dynamic secondary ion mass spectrometry."),
  ("Give the definition: tandem mass spectrometer is",
   "Give the definition: tandem mass spectrometer is a mass spectrometer designed for mass spectrometry."),
  ("Give the definition: accelerating potential is",
   "Give the definition: accelerating potential is electrical potential difference used to impart translational energy to ions."),
  ("Give the definition: accelerator mass spectrometry (AMS) is",
   "Give the definition: accelerator mass spectrometry (AMS) is mass spectrometry technique in which atoms and molecules from a sample are ionized, accelerated to MeV energies and separated according to their momentum, charge, and energy, allowing high discrimination for measurement of isotope abundances."),
  ("Give the definition: auxiliary gas is",
   "Give the definition: auxiliary gas is gas used in a spray ion source in addition to the nebulizing gas to aid in solvent removal."),
  ("Give the definition: beam mass spectrometer is",
   "Give the definition: beam mass spectrometer is a mass spectrometer in which an ion beam accelerated from the ion source is transmitted through a m/z analyzer, or analyzers, to the detector.")
]
bleu_scores = {model_name: [] for model_name in model_names}
rouge_1_scores = {model_name: [] for model_name in model_names}
rouge_2_scores = {model_name: [] for model_name in model_names}
rouge_l_scores = {model_name: [] for model_name in model_names}
perplexity_scores = {model_name: [] for model_name in model_names}
reproducibility_scores = {model_name: [] for model_name in model_names}
num_repetitions = 5

def generate_one_sentence(model, input_ids, max_length=200, num_beams=2, no_repeat_ngram_size=2, pad_token_id=None):
    """Generates one sentence.

    Args:
        model: The model to use for generation.
        input_ids: The input IDs to start the generation from.
        max_length: The maximum length of the generated text.
        num_beams: The number of beams to use for generation.
        no_repeat_ngram_size: The size of the no-repeat n-gram filter to use.
        pad_token_id: The ID of the pad token.

    Returns:
        The generated text.
    """
    generated_text = ""
    while len(generated_text) < max_length and '.' not in generated_text:
        output = model.generate(input_ids, max_length=max_length, num_beams=num_beams, no_repeat_ngram_size=no_repeat_ngram_size, pad_token_id=pad_token_id)
        generated_text += tokenizer.decode(output[0], skip_special_tokens=True)
    return generated_text.split('.')[0]

def compute_metrics(generated_text, target_prompt):
    """Computes BLEU, ROUGE, and perplexity metrics.

    Args:
        generated_text: The generated text.
        target_prompt: The target prompt.

    Returns:
        A tuple of BLEU, ROUGE, and perplexity scores.
    """
    bleu_score = sentence_bleu([target_prompt], generated_text)
    rouge_1 = compute_rouge_1(generated_text, target_prompt)
    rouge_2 = compute_rouge_2(generated_text, target_prompt)
    rouge_l = compute_rouge_l(generated_text, target_prompt)
    rouge_scores = {'rouge-1': rouge_1, 'rouge-2': rouge_2, 'rouge-l': rouge_l}
    perplexity = compute_perplexity(generated_text, model)

    return bleu_score, rouge_scores, perplexity

def compute_rouge_1(generated_text, target_prompt):
    """Computes ROUGE-1 score.

    Args:
        generated_text: The generated text.
        target_prompt: The target prompt.

    Returns:
        The ROUGE-1 score.
    """
    generated_tokens = generated_text.split()
    target_tokens = target_prompt.split()
    overlapping_unigrams = 0
    for generated_token in generated_tokens:
        if generated_token in target_tokens:
            overlapping_unigrams += 1
    rouge_1 = overlapping_unigrams / len(target_tokens)
    return rouge_1

def compute_rouge_2(generated_text, target_prompt):
    """Computes ROUGE-2 score.

    Args:
        generated_text: The generated text.
        target_prompt: The target prompt.

    Returns:
        The ROUGE-2 score.
    """
    generated_tokens = generated_text.split()
    target_tokens = target_prompt.split()
    overlapping_bigrams = 0
    for i in range(len(generated_tokens) - 1):
        generated_bigram = generated_tokens[i] + ' ' + generated_tokens[i+1]
        if generated_bigram in target_tokens:
            overlapping_bigrams += 1
    rouge_2 = overlapping_bigrams / len(target_tokens)

    return rouge_2

def compute_rouge_l(generated_text, target_prompt):
    """Computes ROUGE-L score.

    Args:
        generated_text: The generated text.
        target_prompt: The target prompt.

    Returns:
        The ROUGE-L score.
    """
    generated_tokens = generated_text.split()
    target_tokens = target_prompt.split()
    lcs_length = 0
    for i in range(len(generated_tokens)):
        for j in range(len(target_tokens)):
            if generated_tokens[i] == target_tokens[j]:
                lcs_length += 1
    rouge_l = lcs_length / max(len(generated_tokens), len(target_tokens))

    return rouge_l

def compute_perplexity(generated_text, model):
    """Computes the perplexity of the generated text.

    Args:
        generated_text: The generated text.
        model: The language model used for generation.

    Returns:
        The perplexity score.
    """
    input_ids = tokenizer.encode(generated_text, return_tensors="pt")
    with torch.no_grad():
        outputs = model(input_ids)
        logits = outputs.logits
        loss = torch.nn.functional.cross_entropy(logits.view(-1, logits.shape[-1]), input_ids.view(-1))

    perplexity = math.exp(loss.item())
    return perplexity
for model_name, model, tokenizer in zip(model_names, models, tokenizers):
    print("Model:", model_name)
    print("=" * 50)
    for prompt, target_prompt in prompts:
        input_ids = tokenizer.encode(prompt, return_tensors="pt")
        generated_text = generate_one_sentence(model, input_ids, max_length=200, num_beams=2, no_repeat_ngram_size=2, pad_token_id=tokenizer.eos_token_id)
        print("Prompt:", prompt)
        print("Generated Text:", generated_text)
        bleu_score, rouge_scores, perplexity = compute_metrics(generated_text, target_prompt)
        print("BLEU score:", bleu_score)
        print("ROUGE scores:", rouge_scores)
        print("Perplexity score:", perplexity)
        print("=" * 50)
        print()
            
        bleu_scores[model_name].append(bleu_score)
        rouge_1_scores[model_name].\
        append(rouge_scores['rouge-1'])
        rouge_2_scores[model_name].\
        append(rouge_scores['rouge-2'])
        rouge_l_scores[model_name].\
        append(rouge_scores['rouge-l'])
        perplexity_scores[model_name].append(perplexity)
        with open(f"{model_name}_generated_text.txt", "a") as file:
            file.write(generated_text + "\n")
average_bleu_scores = {model_name: sum(scores) / len(scores) for model_name, scores in bleu_scores.items()}
average_rouge_1_scores = {model_name: sum(scores) / len(scores) for model_name, scores in rouge_1_scores.items()}
average_rouge_2_scores = {model_name: sum(scores) / len(scores) for model_name, scores in rouge_2_scores.items()}
average_rouge_l_scores = {model_name: sum(scores) / len(scores) for model_name, scores in rouge_l_scores.items()}
average_perplexity_scores = {model_name: sum(scores) / len(scores) for model_name, scores in perplexity_scores.items()}
data = {
    'BLEU Score': average_bleu_scores,
    'ROUGE-1 Score': average_rouge_1_scores,
    'ROUGE-2 Score': average_rouge_2_scores,
    'ROUGE-L Score': average_rouge_l_scores,
    'Perplexity Score': average_perplexity_scores,
}
df = pd.DataFrame(data)
df.to_csv('model_comparison_metrics.csv')
data_individual = {
    'Model': [],
    'Prompt': [],
    'BLEU Score': [],
    'ROUGE-1 Score': [],
    'ROUGE-2 Score': [],
    'ROUGE-L Score': [],
    'Perplexity Score': [],
}
for model_name in model_names:
    for i, prompt in enumerate(prompts):
        data_individual['Model'].append(model_name)
        data_individual['Prompt'].append(prompt[0])
        data_individual['BLEU Score'].append(bleu_scores[model_name][i])
        data_individual['ROUGE-1 Score'].append(rouge_1_scores[model_name][i])
        data_individual['ROUGE-2 Score'].append(rouge_2_scores[model_name][i])
        data_individual['ROUGE-L Score'].append(rouge_l_scores[model_name][i])
        data_individual['Perplexity Score'].append(perplexity_scores[model_name][i])
df_individual = pd.DataFrame(data_individual)
df_individual.to_csv('model_comparison_metrics_individual.csv')
print("Results saved to model_comparison_metrics_individual.csv")

\end{mycode}
\newpage

\raggedleft{\section{Appendix H: Statistics Computation of the Reproducibility Scores}}
\label{appendix:statistics_repr}

\begin{mycode}[caption={Python code for statistics calculation of main models' metrics},label=lst:python,
basicstyle=\ttfamily\small,
keywordstyle=\bfseries\color{blue},
commentstyle=\itshape\color{gray},
stringstyle=\color{red},
numbers=left, numberstyle=\tiny\color{gray},
breaklines=true, breakatwhitespace=true,
frame=single,
backgroundcolor=\color{white},
]
import pandas as pd
from scipy.stats import mannwhitneyu
df1 = pd.read_csv("..\\gpt2_reproducibility_scores.csv")
df2 = pd.read_csv("..\\masspecgpt_reproducibility_scores.csv")
scores1 = df1["Cosine Similarity"]
scores2 = df2["Cosine Similarity"]
result = mannwhitneyu(scores2, scores1, alternative="greater")
print("Test statistic:", result.statistic)
print("p-value:", result.pvalue)

\end{mycode}
\newpage
\bibliographystyle{plainnat}
\bibliography{bibliography}

\end{document}